\documentclass[letterpaper]{article} 
\usepackage{aaai24}  
\usepackage{times}  
\usepackage{helvet}  
\usepackage{courier}  
\usepackage[hyphens]{url}  
\usepackage{graphicx} 
\usepackage{natbib}  
\usepackage{caption} 
\usepackage{kotex}
\usepackage{algorithm}
\usepackage{algorithmic}

\usepackage{amsmath,amssymb} 
\usepackage{multirow}
\usepackage[table,xcdraw]{xcolor}
\usepackage{subcaption}
\usepackage{newfloat}
\usepackage{listings}
\DeclareCaptionStyle{ruled}{labelfont=normalfont,labelsep=colon,strut=off} 
\floatstyle{ruled}
\newfloat{listing}{tb}{lst}{}
\floatname{listing}{Listing}
\usepackage{booktabs}

\usepackage{lineno}

\title{U-MixFormer: UNet-like Transformer with Mix-Attention \\ for Efficient Semantic Segmentation}
\author{
    Seul-Ki Yeom\thanks{Corresponding author.}, Julian von Klitzing
}
\affiliations{
    Nota AI GmbH\\
    Friedrichstrasse 200, 10117 Berlin, Germany\\
    skyeom@nota.ai, julian.von.klitzing@campus.tu-berlin.de
}

\begin{document}

\maketitle

\begin{abstract}
Semantic segmentation has witnessed remarkable advancements with the adaptation of the Transformer architecture. Parallel to the strides made by the Transformer, CNN-based U-Net has seen significant progress, especially in high-resolution medical imaging and remote sensing. This dual success inspired us to merge the strengths of both, leading to the inception of a U-Net-based vision transformer decoder tailored for efficient contextual encoding.
Here, we propose a novel transformer decoder, U-MixFormer, built upon the U-Net structure, designed for efficient semantic segmentation. Our approach distinguishes itself from the previous transformer methods by leveraging lateral connections between encoder and decoder stages as feature queries for the attention modules, apart from the traditional reliance on skip connections.
Moreover, we innovatively mix hierarchical feature maps from various encoder and decoder stages to form a unified representation for keys and values, giving rise to our unique \emph{mix-attention} module.

Our approach demonstrates state-of-the-art performance across various configurations. Extensive experiments show that U-MixFormer outperforms SegFormer, FeedFormer, and SegNeXt by a large margin. For example, U-MixFormer-B0 surpasses SegFormer-B0 and FeedFormer-B0 with 3.8\% and 2.0\% higher mIoU and 27.3\% and 21.8\% less computation and outperforms SegNext with 3.3\% higher mIoU with MSCAN-T encoder on ADE20K.
Code available at \url{https://github.com/julian-klitzing/u-mixformer}.

\end{abstract}

\section{Introduction}
\label{sec:intro}
    
\begin{figure}[t]
\centering
\includegraphics[width=1.0\columnwidth]{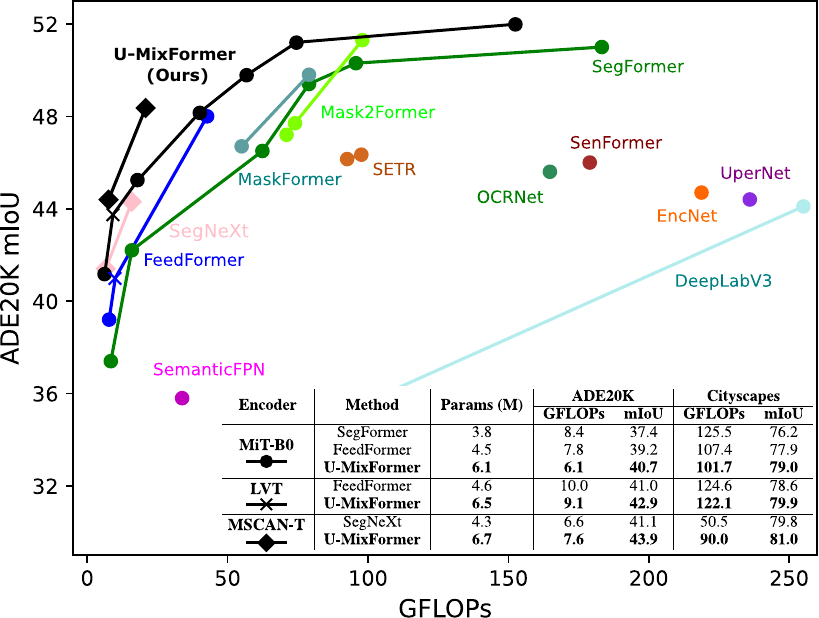}
\caption{Performance vs. computational efficiency on ADE20K (single-scale inference). U-MixFormer outperforms previous methods in all configurations.}
\label{fig:fig1_performance_comparison}
\end{figure}

Semantic segmentation, a fundamental downstream task in computer vision, has consistently received increasing attention in industry and academia. 
The importance of semantic segmentation is highlighted by its widespread applications in real-world scenarios such as autonomous driving~\cite{WangLLPTS22} and medical diagnosis~\cite{ChaitanyaEKK20}. Despite these advancements, achieving precise pixel-wise predictions remains a challenge due to the need to balance global and local contexts.

The introduction of the fully convolutional network (FCN)\cite{FCN} popularized the encoder-decoder structure, in which the encoder extracts high-level semantics and the decoder combines them with spatial details. While variants\cite{SegNet, DeepLab} have improved this approach, traditional CNNs struggle to capture long-range context. This limitation has prompted interest in vision transformer-based methods for segmentation.

Transformer~\cite{AttentionIAYN}, initially designed for natural language processing, has been adapted for vision tasks with notable success thanks to the self-attention mechanism to capture global relationships within input sequences. Inspired by its success, \citet{ViT} adapted it for vision tasks, leading to the Vision Transformer (ViT), which interprets images as sequences of embedded patches and processes them using the Transformer encoder. This approach yielded remarkable results on ImageNet. Since the introduction of the ViT, numerous studies have been performed into its adaptation for semantic segmentation. The primary objectives have been two-fold: refining the encoder and crafting decoders that adeptly utilize features from the encoder stage.

There has been a notable shift towards utilizing the transformer's decoder structure in vision tasks. 
DETR by~\citet{DETR} pioneered this approach and integrated the transformer encoder-decoder framework into detection and segmentation. Following DETR, Segmenter~\cite{Segmentor}, MaskFormer~\cite{MaskFormer}, and Mask2Former~\cite{Mask2Former} introduced decoders for mask prediction with global class labels, emphasizing high-level features. More recently, FeedFormer~\cite{FeedFormer} proposed a decoder design that decodes high-level encoder features using only the lowest-level encoder feature. Despite the ongoing progress on transformer-based decoders for segmentation, these methods often rely on computationally intensive feature configurations within their attention mechanisms. Moreover, these methods exhibit inefficiencies in the propagation of feature maps across the decoder stages.


Traditionally, the U-Net architecture~\cite{UNet}, characterized by its symmetric CNN-based encoder-decoder structure, has been a favored choice for semantic segmentation, particularly in the medical field. This favor stems from U-Net's characteristics of effectively capturing and propagating hierarchical features. Furthermore, its lateral connections play an important role, facilitating the flow of multi-stage features between the encoder and decoder. We hypothesize that leveraging these inherent strengths of the U-Net architecture can lead to efficient refinement of features, which can then be hierarchically integrated into the transformer decoder phase.

In this paper, we propose \emph{U-MixFormer}, a novel UNet-like transformer decoder.
Building upon the foundational principles of U-Net, U-MixFormer adaptively integrates multi-stage features as keys and values within its dedicated \emph{mix-attention} module. This module ensures the gradual propagation of features and successively remixes them across decoder stages, effectively managing dependencies between these stages to capture context and refine boundaries. This can emphasize the hierarchical feature representation like traditional CNNs and enhance the global context understanding capabilities of Transformers. To the best of our knowledge, it is the first work to synergize the inherent strengths of U-Net with the transformative capabilities of Vision Transformers, particularly through a novel attention module for effectively harmonizing queries, keys, and values for semantic segmentation.


Our contributions are summarized as follows:
\begin{itemize}
    \item \textbf{Novel Decoder Architecture with U-Net} We propose a novel powerful transformer-decoder architecture motivated by the U-Net for efficient semantic segmentation. Capitalizing on U-Net's proficiency in capturing and propagating hierarchical features, our design distinctively uses the lateral connections of a transformer encoder as query features. This approach ensures a harmonious fusion of high-level semantics and low-level structures.
    \item \textbf{Optimized Feature Synthesis for Enhanced Contextual Understanding} To improve the efficiency of our UNet-like transformer architecture, we mix and update multiple encoder and decoder outputs as integrated features for keys and values, resulting in our proposed \emph{mix-attention} mechanism.
    This approach has not only rich feature representation for each decoder stage but also boosts contextual understanding.
    \item \textbf{Compatibility with Diverse Encoders} We demonstrate the compatibility of U-MixFormer combined with the existing popular encoders of both transformer-based (MiT and LVT) and CNN-based (MSCAN) encoders.
    \item \textbf{Empirical Benchmarking} As shown in Figure~\ref{fig:fig1_performance_comparison}, U-MixFormer achieves a new state-of-the-art in terms of computational cost as well as accuracy among semantic segmentation methods. It consistently outperforms light-weight, middle-weight, and even heavy-weight encoders. This superiority is demonstrated for the ADE20K and Cityscapes datasets, with notable performance on the challenging Cityscape-C dataset.
\end{itemize}

\begin{figure*}[h]
\centering
\includegraphics[width=0.8\textwidth]{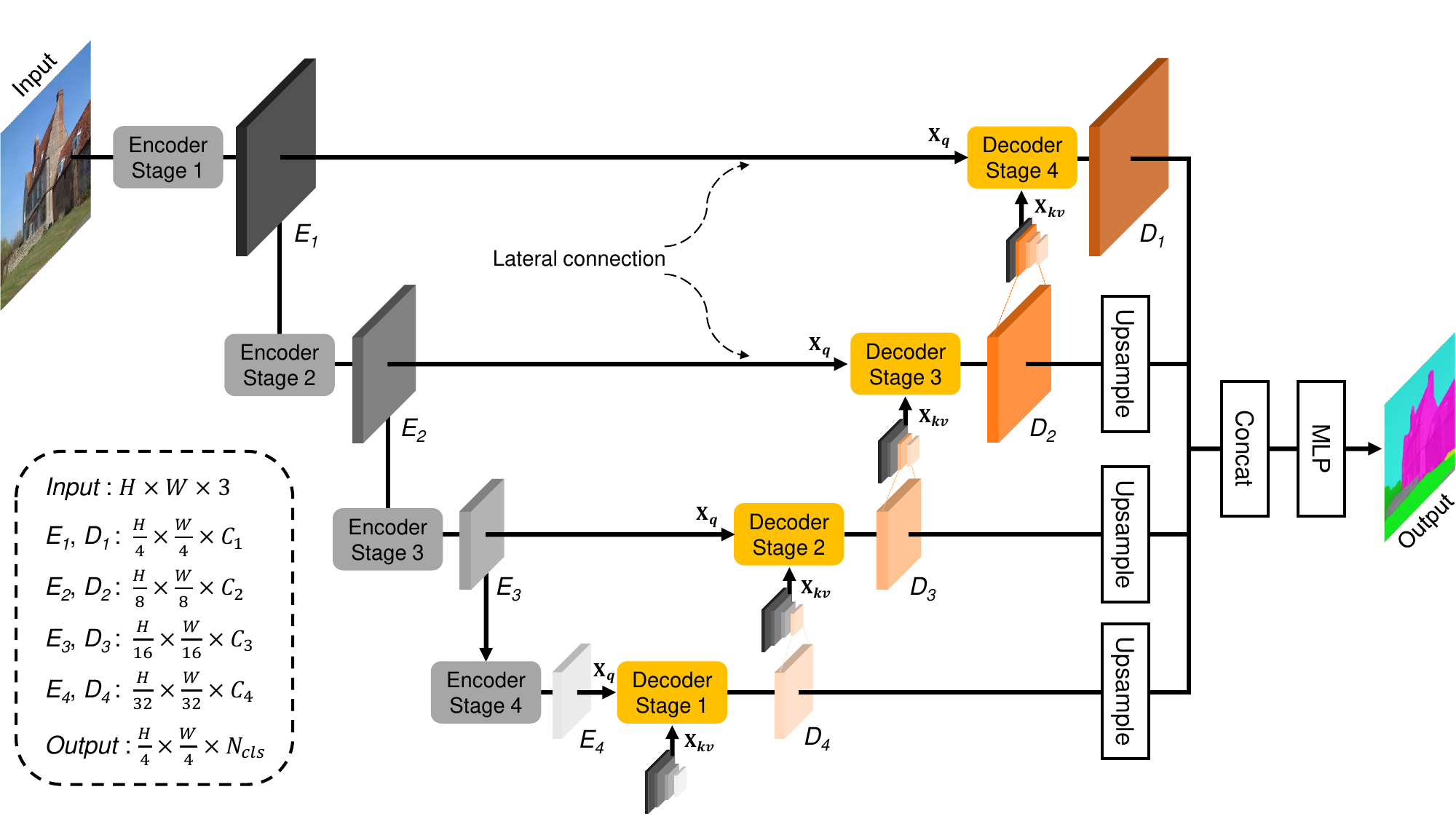}
\caption{U-MixFormer architecture: Encoder (left) extracts multi-resolution feature maps from input image. U-MixFormer decoder (right) fuses lateral encoder outputs as $\mathbf{X_q}$ with previous decoder stage outputs by incorporating them into $\mathbf{X_{kv}}$ using our mixed-attention mechanism. Finally, feature maps from all decoder stages are concatenated, and a MLP predicts the output.}
\label{fig:model_architecture}
\end{figure*}
\section{Related Work}
\label{sec:relatedwork}
\subsection{Encoder Architectures}
SETR \cite{SETR} was the first architecture to adopt ViT as an encoder for semantic segmentation. Because ViT only divides the input image into patches, SETR produces single-scaled encoder features. PVT \cite{PVT} and Swin Transformer \cite{SwinTrafo} repeatedly group feature maps into new non-overlapping patches between encoder stages, thereby hierarchically generating multi-scale encoder features.
Both methods also enhance the efficiency of the self-attention module by either reducing the spatial dimensions of keys and values (PVT) or grouping patches with shifted windows (Swin Transformer). SegFormer \cite{SegFormer} reuses PVT's efficiency strategy while removing positional encodings and embedding feature maps into overlapping patches.
In contrast to the previously mentioned methods, the encoders of SegNeXt \cite{SegNeXt} and LVT \cite{LVT} employs convolutional attention mechanisms.

\subsection{Decoder Architectures}
DETR \cite{DETR} was the first method to deploy a transformer decoder for semantic segmentation. Subsequent works \cite{Segmentor, MaskFormer, Mask2Former} adapted DETR but also rely on object-learnable queries, which are computationally expensive, particularly when combined with multi-scale encoder features. In contrast, FeedFormer \cite{FeedFormer} directly utilizes the features from the encoder stages as feature queries, leading to enhanced efficiency. FeedFormer decodes the high-level encoder features (used as features for queries) with the lowest-level encoder feature (used as features for keys and values). However, this setup processes the feature maps individually, without incremental propagation of feature maps between decoder stages, thereby missing the opportunity for more incremental refinement to improve object boundary detection. Additionally, other recent MLP or CNN-based decoders \cite{SegFormer, SegNeXt} also lack incremental propagation of decoder features.

\subsection{UNet-like Transformer}
In both the medical and remote sensing domains, efforts have been made to transition the UNet architecture from a CNN-based framework to a transformer-based one. TransUNet \cite{TransUNet} marked the first successful endeavor to incorporate the Transformer into medical image segmentation, using ViT in conjunction with their CNN encoder. Other hybrid approaches are presented in \cite{UTransformer, UNETR, UNetFormer}. \citet{SwinUNetMedical} introduced Swin-UNet, the first fully transformer-based UNet-like architecture. This design features heavyweight Swin Transformer stages for both the encoder and decoder, preserving the lateral connections between them as skip connections. \emph{In contrast to Swin-UNet, our approach employs lighter-weight decoder stages, rendering it suitable for a wider range of downstream tasks. Furthermore, we interpret the lateral connections as features for queries instead of as skip connections and incorporate a unique attention mechanism.}
\section{Proposed Method}
\label{sec:method}
This section introduces U-MixFormer, a novel UNet-like transformer decoder architecture for semantic segmentation.
In general, our decoder is composed of as many stages $i\in\{1, ..., N\}$ as there are encoder stages. For clarity, Figure~\ref{fig:model_architecture} provides a visual overview of this architecture, exemplified with a four-stage ($N = 4$) hierarchical encoder such as MiT, LVT, or MSCAN. 

First, the encoder processes an input image with $H \times W \times 3$. The four stages  $i\in\{1, ..., 4\}$ yield hierarchical, multi-resolution features $\mathbf{E}_i$ with $\frac{H}{2^{i+1}} \times \frac{W}{2^{i+1}} \times C_i$. 
Second, our decoder stages $i$ sequentially generate refined features $\mathbf{D}_{4 - i +1}$ by performing \emph{mix-attention} where features for queries $\mathbf{X}_{q}^i$ equal the respective lateral encoder feature map. The features for keys and values $\mathbf{X}_{kv}^i$ are given by a mix of encoder and decoder stages. Notably, our decoder mirrors the dimensions of the encoder stage outputs.
Third, the decoder features are upsampled using bilinear interpolation to match the height and width of $\mathbf{D}_1$. Finally, the concatenated features are processed by an MLP to predict the segmentation map with $H/4 \times W/4 \times 3$.

\subsection{Mix-Attention}
Attention modules used in transformer blocks compute the scaled dot-product attention for queries $\mathbf{Q}$, keys $\mathbf{K}$ and values $\mathbf{V}$ as follows: 
\begin{equation}
    \label{eq_attention}
    Attention(\mathbf{Q}, \mathbf{K}, \mathbf{V}) = Softmax(\frac{\mathbf{Q}\mathbf{K}^\top}{\sqrt{d_k}}) \mathbf{V}
\end{equation}
where $d_k$ is the embedding dimension of the keys and $\mathbf{Q}$, $\mathbf{K}$, $\mathbf{V}$ are obtained from linear projections of selected features.

\begin{figure}[h]
\centering
\includegraphics[width=0.80\columnwidth]{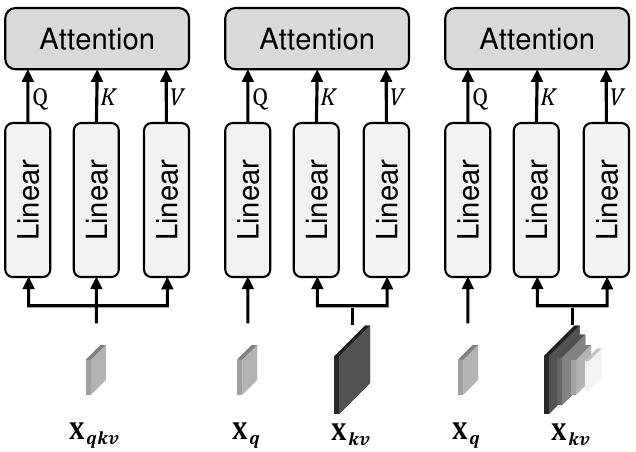}
\caption{Comparison of three different atttention module: Self- (left), cross- (middle) and mix-attention (right).}
\label{fig:attention}
\end{figure}

Central to our method is the selection of features $\mathbf{X_{kv}}$ that are to be projected to keys and values, which results in our proposed mix-attention mechanism. A comparison between traditional self-, cross-, and the novel \emph{mix-attention} is illustrated in Figure~\ref{fig:attention}.

In self-attention, the features used to generate queries, keys and values are identical ($\mathbf{X_{qkv}}$) and originate from the same source, i.e. the same encoder/decoder stage. Cross-attention employs two distinct features, $\mathbf{X_q}$ and $\mathbf{X_{kv}}$, each derived from a \emph{single} unique source. 
In contrast, our mix-attention mechanism utilizes a mixed feature for $\mathbf{X_{kv}}$ sourced from \emph{multiple} multi-scale stages. This concept allows the query to find matches across all different stages, i.e. degrees of contextual granularity, thereby facilitating enhanced feature refinement. The efficacy of this method is validated by our experiments in the Ablation Studies section.


The selection of a feature set $\mathcal{F}^i$ for a decoder stage $i$ is formalized piecewise as follows: 
\begin{equation}
\label{eq_selection_kv}
    \mathcal{F}^i = \left\{
	\begin{array}{cc}
		\{\mathbf{E}_j\}_{j=1}^{N}  & \text{if } i = 1 \\[5pt]
		\{\mathbf{E}_j\}_{j=1}^{N-i+1} \cup \{\mathbf{D}_j\}_{j=N-i+2}^{N} & \text{otherwise}
	\end{array}
\right.
\end{equation}
where for the first decoder stage ($i = 1$) all encoder features are selected. For subsequent stages, previously computed decoder stage outputs are propagated by replacing their lateral encoder counterparts in $\mathcal{F}^i$. 

To align the spatial dimensions of the features in $\mathcal{F}^i$, we adapt the \emph{spatial reduction} procedure introduced by \citet{PVT} : 
\begin{equation}
\begin{aligned}
\label{eq_spatial_reduction}
    \mathbf{\hat{F}}_j^i &= AvgPool(pr_j, pr_j)(\mathcal{F}^i_j), \forall j \in \{1, ..., N-1\} \\
    \mathbf{\hat{F}}_j^i &= Linear(C_{j}, C_{j})(\mathbf{\hat{F}}_j^i), \forall j \in \{1, ..., N-1\}
\end{aligned}
\end{equation}
where $\mathcal{F}^i_j$ denotes the $j$-th element of feature set $\mathcal{F}^i$, and $pr_j$ is the pooling ratio which aligns the size with the smallest feature map $\mathcal{F}^i_N$. The operations AvgPool and Linear are configured as AvgPool($kernelSize$, $stride$)($\cdot$) and Linear($C_{in}$, $C_{out}$)($\cdot$), respectively.

The spatially aligned features are concatenated along the channel dimension to form a mixed feature $\mathbf{X}_{kv}^i$ for keys and values.

\begin{equation}
\label{eq_concat}
    \mathbf{X}_{kv}^{i} = Concat(\{\mathbf{\hat{F}}_j^i\}_{j=1}^{N-1} \cup \{\mathcal{F}_{N}^{i}\})  
\end{equation}
 
\subsection{Decoder Stage}
\begin{figure}[thb]
\centering
\includegraphics[width=0.80\columnwidth]{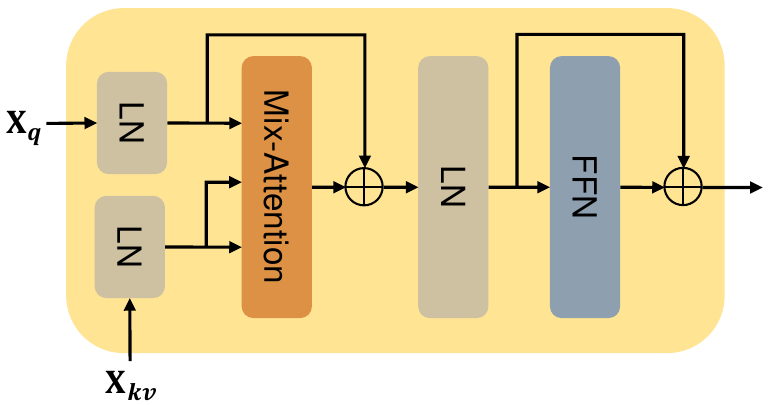}
\caption{Structure of decoder stage employing our mix-attention mechanism and taking features for queries $\mathbf{X}_{q}$ and a mixed feature for keys and values $\mathbf{X}_{kv}$ as input.}
\label{fig:decoder_block}
\end{figure}

We adapt the traditional transformer decoder block, by discarding the self-attention module as recommended by~\citet{FeedFormer}. Additionally, we replace the cross-attention module with our proposed mix-attention module. The resulting structure is depicted in Figure \ref{fig:decoder_block}.

Using layer normalization (LN) and a feedforward network (FFN), the output for a $DecoderStage_i$ is computed as follows:
\begin{equation}
\begin{aligned}
   \mathbf{A}_i = LN(MixAtt.(LN(\mathbf{X}_{kv}^{i}, \mathbf{X}_{q}^{i})) + LN(\mathbf{X}_{q}^{i})) \\
    DecoderStage_i = \mathbf{D}_{N - i +1} = FFN(\mathbf{A}_i) + \mathbf{A}_i
\end{aligned}
\end{equation}
where $MixAtt.$ denotes our proposed mix-attention.

\begin{table*}[]
    \centering
    \caption{Performance comparison with the state-of-the art methods on ADE20K and Cityscapes.}
    \label{tab:perform_comparison_lightmiddle}
    \scalebox{0.80}{
\begin{tabular}{c|l|c|c|cc|cc}
\specialrule{1pt}{1pt}{1pt}
\hline
                                          &                                   &                                    &                                       & \multicolumn{2}{c|}{\textbf{ADE20K}}       & \multicolumn{2}{c}{\textbf{Cityscapes}}    \\ \cline{5-8} 
\multirow{-2}{*}{\textbf{}}               & \multirow{-2}{*}{\textbf{Method}} & \multirow{-2}{*}{\textbf{Encoder}} & \multirow{-2}{*}{\textbf{Params (M)}~$\downarrow$} & \textbf{GFLOPs}~$\downarrow$ & \textbf{mIoU}~$\uparrow$            & \textbf{GFLOPs}~$\downarrow$ & \textbf{mIoU}~$\uparrow$            \\ \hline
                                          & FCN~\cite{FCN}                               & MobileNetV2                        & 9.8                                   & 39.6            & 19.7                     & 317.1           & 61.5                     \\
                                          & PSPNet~\cite{zhao2017PSNet}                            & MobileNetV2                        & 13.7                                  & 52.9            & 29.6                     & 423.4           & 70.2                     \\
                                          & DeepLabV3+~\cite{DeepLab}                        & MobileNetV2                        & 15.4                                  & 69.4            & 34.0                     & 125.5           & 75.2                     \\
                                          & SwiftNetRN~\cite{OrsicKBS19}                        & ResNet-18                          & 11.8                                  & -               & -                        & 104.0           & 75.4                     \\
                                          & Semantic FPN~\cite{Li21ConvMLP}                      & ConvMLP-S                          & 12.8                                  & 33.8            & 35.8                     & -               & -                        \\
                                          & SegFormer~\cite{SegFormer}                         & MiT-B0                             & 3.8                                   & 8.4             & 37.4                     & 125.5           & 76.2                     \\
                                          & FeedFormer~\cite{FeedFormer}                        & MiT-B0                             & 4.5                                   & 7.8             & 39.2                     & 107.4           & 77.9                     \\
                                        & FeedFormer~\cite{FeedFormer}                        & LVT                                & 4.6                                   & 10.0            & 41.0                     & 124.6           & 78.6                     \\
                                          & SegNeXt~\cite{SegNeXt}                           & MSCAN-T                            & 4.3                                   & 6.6             & 41.1                     & 50.5            & 79.8                     \\
                                          & \textbf{U-MixFormer (Ours)}            & \textbf{MiT-B0}                    & \textbf{6.1}                          & \textbf{6.1}    & \textbf{41.2}            & \textbf{101.7}  & \textbf{79.0}            \\
                                          & \textbf{U-MixFormer (Ours)}            & \textbf{LVT}                       & \textbf{6.5}                          & \textbf{9.1}    & \textbf{43.7}            & \textbf{122.1}  & \textbf{79.9}            \\
\multirow{-12}{*}{\rotatebox{90}{\textbf{Light-weight}}}  & \textbf{U-MixFormer (Ours)}            & \textbf{MSCAN-T}                   & \textbf{6.7}                          & \textbf{7.6}    & \textbf{44.4}            & \textbf{90.0}   & \textbf{81.0}            \\ \hline
                                          & CCNet~\cite{Huang19CCNet}                             & ResNet-101                         & 68.9                                  & 278.4           & 43.7                     & 2224.8          & 79.5                     \\
                                          & DeepLabV3+~\cite{DeepLab}                        & ResNet-101                         & 52.7                                  & 255.1           & 44.1                     & 2032.3          & 80.9                     \\
                                          & Auto-DeepLab~\cite{Liu19AutoDeepLab}                      & Auto-DeepLab-L                     & 44.4                                  & -               & -                        & 695.0           & 80.3                     \\
                                          & OCRNet~\cite{Yuan20OCRNet}                            & HRNet-W48                          & 70.5                                  & 164.8           & 45.6                     & 1296.8          & 81.1                     \\
                                          & Seg-S-/16~\cite{Segmentor}                         & ViT-S                              & 22.0                                  & 31.8            & 45.4                     & -               & -                        \\
                                          & SegFormer~\cite{SegFormer}                         & MiT-B2                             & 27.5                                  & 62.4            & 46.5                     & 717.1           & 81.0                     \\
                                          & MaskFormer~\cite{MaskFormer}                        & Swin-T                             & 42.0                                  & 55.0            & 46.7                     & -               & -                        \\
                                          & Mask2Former~\cite{Mask2Former}                       & Swin-T                             & 47.0                                  & 74.0            & 47.7                     & -               & 82.1                     \\
                                          & Mask2Former~\cite{Mask2Former}                       & ResNet-101                         & 63.0                                  & 90.0            & 47.8                     & -               & -                        \\
                                          & SegFormer~\cite{SegFormer}                         & MiT-B1                             & 13.7                                  & 15.9            & 42.2                     & 243.7           & 78.5                     \\
                                        & SegNeXt~\cite{SegNeXt}                           & MSCAN-S                            & 13.9                                  & 15.9            & 44.3                     & 124.6           & 81.3                     \\
                                          & \textbf{U-MixFormer (Ours)}            & \textbf{MiT-B1}                    & \textbf{24.0}                         & \textbf{17.8}   & \textbf{45.2}            & \textbf{246.8}  & \textbf{79.9} \\
                                        & SegFormer~\cite{SegFormer}                        & MiT-B2                             & 27.5                                  & 62.4            & 46.5                     & 717.1           & 81.0                     \\                                          
                                        & FeedFormer~\cite{FeedFormer}                        & MiT-B2                             & 29.1                                  & 42.7            & 48.0                     & 522.7           & 81.5                     \\
                                         & \textbf{U-MixFormer (Ours)}            & \textbf{MiT-B2}                    & \textbf{35.8}                         & \textbf{40.0}   & \textbf{48.2}            & \textbf{515.0}  & \textbf{81.7}            \\
\multirow{-16}{*}{\rotatebox{90}{\textbf{Middle-weight}}} & \textbf{U-MixFormer-S (Ours)}            & \textbf{MSCAN-S}                   & \textbf{24.3}                         & \textbf{20.8}   & \textbf{48.4} & \textbf{154.0}  & \textbf{81.8} \\
 \hline
\specialrule{1pt}{1pt}{1pt}
\end{tabular}
    }
\end{table*}

\begin{table}[]
    \centering
    \caption{Performance comparison for heavyweight encoders on ADE20K.}
    \label{append_tab:perform_comparison_heavy}
    \scalebox{0.75}{
\begin{tabular}{c|c|c|cc}
\specialrule{1pt}{1pt}{1pt}
\hline
\multicolumn{1}{c|}{\textbf{Method}} & \multicolumn{1}{c|}{\textbf{Encoder}} & \multicolumn{1}{c|}{\textbf{Params (M)}~$\downarrow$} & \multicolumn{1}{c}{\textbf{GFLOPs}~$\downarrow$} & \multicolumn{1}{c}{\textbf{mIoU}~$\uparrow$}            \\ \hline
Seg-B-Mask/16                        & ViT-B                                 & 102.4                                    & 129.4                               & 48.7                                         \\
Seg-L-Mask/16                        & ViT-L                                 & 333.2                                    & 399.9                               & 51.8                                         \\
SETR-PUP                             & ViT-L                                 & 318.3                                    & 425.9                               & 48.6                                         \\
SETR-MLA                             & ViT-L                                 & 310.5                                    & 368.4                               & 48.6                                         \\
MaskFormer                           & Swin-S                                & 63.0                                     & 79.0                                & 49.8                                         \\
SegFormer                            & MiT-B3                                & 47.3                                     & 79.0                                & 49.4                                         \\
\textbf{U-MixFormer  (Ours)}               & \textbf{MiT-B3}                       & \textbf{55.7}                            & \textbf{56.8}                       & \textbf{49.8}                                \\
SegFormer                   & MiT-B4                       & 64.1                            & 95.7                       & 50.3                             \\
\textbf{U-MixFormer  (Ours)}               & \textbf{MiT-B4}                       & \textbf{72.4}                            & \textbf{73.4}                       & \textbf{50.4}                                \\
SegFormer                            & MiT-B5                                & 84.7                                     & 183.3                               & 51.0                                         \\
\textbf{U-MixFormer+  (Ours)}              & \textbf{MiT-B4}                       & \textbf{72.4}                            & \textbf{74.5}                       & \textbf{51.2}                                \\
\textbf{U-MixFormer  (Ours)}               & \textbf{MiT-B5}                       & \textbf{93.0}                            & \textbf{149.5}                      & \textbf{51.9}                                \\
\textbf{U-MixFormer+  (Ours)}              & \textbf{MiT-B5}                       & \textbf{93.0}                            & \textbf{152.0}                      & \textbf{52.0} \\ \hline
\specialrule{1pt}{1pt}{1pt}
\end{tabular}
}
\end{table}

\subsection{Relationship to UNet Architectures}

We introduced U-MixFormer as a UNet-like architecture. However, we want to underline the main differences between our approach and other UNet-like variants.

\begin{itemize}
    \item Because we view the lateral connections as features for queries, our decoder feature maps implicitly increase in spatial resolution without the need for explicit upsampling between decoder stages.
    \item Our approach uses all decoder stages to predict the segmentation map, not only the final one.  
    \item The feature map of the last decoder stage yields a resolution of $H/4$, $W/4$, whereas others restore the original spatial resolution $H$, $W$.
\end{itemize}

\section{Experiments}
\label{sec:experiments}
\subsection{Experimental Settings}
\subsubsection{Datasets}
Experiments are conducted on two popular benchmark datasets: ADE20K~\cite{Zhou17ADE} and Cityscapes~\cite{Cordts16Cityscape}. ADE20K is a rigorous scene parsing benchmark highlighting 150 intricate semantic concepts, split into 20,210 images for training and 2,000 for validation. Cityscapes incorporates 19 densely annotated object categories from urban imagery, aggregating 5,000 images of a high resolution of 2048~$\times$~1024. It also introduces 19,998 roughly annotated images for enhanced model training.

\begin{figure*}[h]
\centering
\includegraphics[width=1.4\columnwidth]{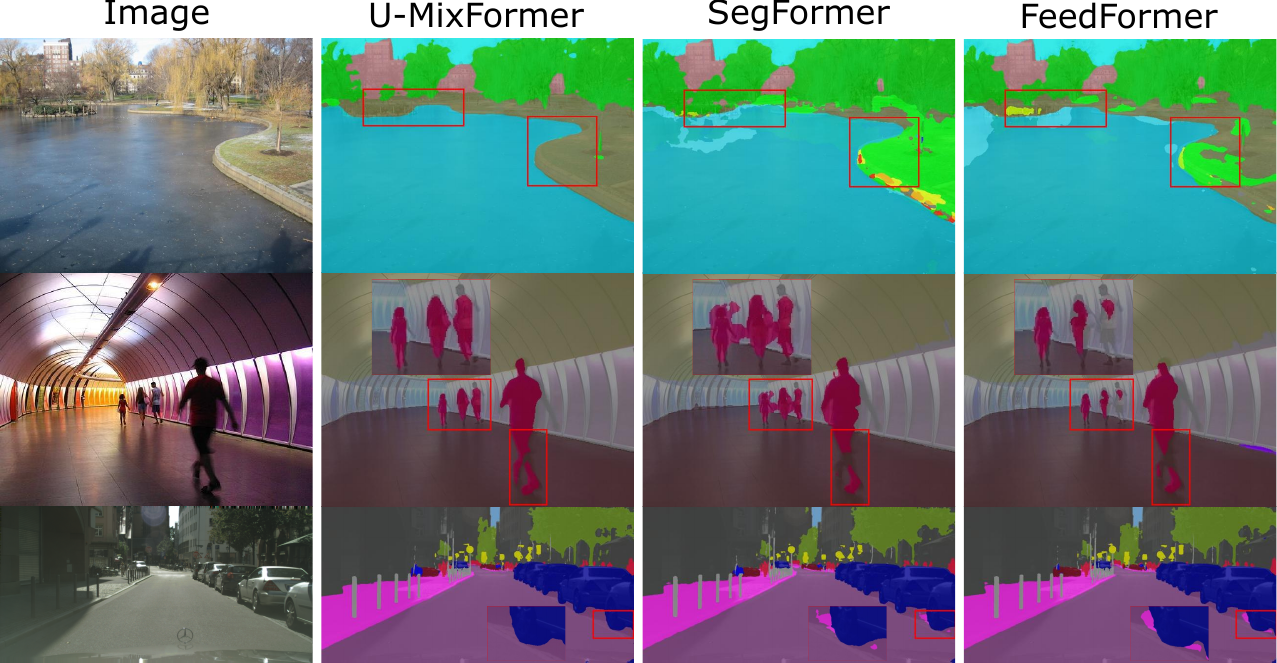}
\caption{Qualitative analysis on ADE20K and Cityscapes datasets: We select SegFormer and FeedFormer as benchmarks since they share the same encoder. Our observations indicate that U-MixFormer outperforms these methods, particularly in segmenting complex object details such as the boundaries between the lake and floor and across human objects.}
\label{fig:qualitative_result}
\end{figure*}
 
\subsubsection{Implementation Details} 
To evaluate the versatility of U-MixFormer across various encoder complexities, we incorporated three distinct encoder backbones: the Mix Transformer (MiT)~\cite{SegFormer}, Light Vision Transformer (LVT)~\cite{LVT}, and the multi-scale convolutional attention-based encoder (MSCAN)~\cite{SegNeXt}. In detail, we utilized MiT-B0, LVT, and MSCAN-T for our light-weight variants and MiT-B1/2 and MSCAN-S for the middle-weight architectures, while the heavier variants included MiT-B3/4/5. The embedding dimensions were 128 in the final MLP phase for light-weight models and 768 for others. Section A.1 of the supplementary material provides additional information about the training and evaluation setting.

\begin{table}[]
    \centering
    \caption{Ablation study showing the effectiveness of mix-attention and U-Net like architecture.}
    \label{tab:ablation_mix_unet}
    \scalebox{0.80}{
\begin{tabular}{c|c|c|l}
\multicolumn{1}{c|}{\textbf{Methods}}             & \textbf{Params (M)} & \textbf{GFLOPs} & \multicolumn{1}{c}{\textbf{mIoU}}        \\ \hline
\multirow{2}{*}{\begin{tabular}[c]{@{}c@{}}Baseline - FeedFormer\\ (Cross-Attention)\end{tabular}} & \multirow{2}{*}{4.5}   & \multirow{2}{*}{7.8}   & \multirow{2}{*}{39.2}           \\
                                                                                             &                        &                        &                                 \\ \cline{1-1}
Mix-Attention                                                                                      & 6.0                  & 5.7                  & \multicolumn{1}{c}{39.9 (+0.7)} \\ \cline{1-1}
\multirow{2}{*}{Cross-Attention + U-Net}  & \multirow{2}{*}{5.0} & \multirow{2}{*}{6.1} & \multirow{2}{*}{40.8 (+0.9)}    \\

                                                                                             &                        &                        &                                 \\ \hline
\begin{tabular}[c]{@{}c@{}}\textbf{Mix-Attention + U-Net} \\ \textbf{(proposed method)}\end{tabular}                 & \textbf{6.1}      & \textbf{6.1}  & \textbf{41.2 (+0.4)}
\end{tabular}
}
\end{table}

\begin{table*}[]
    \centering
    \caption{The average mIoU values for both the clean and corrupted variants of the Cityscapes validation set were computed for three semantic segmentation methods, all of which utilize the same encoder (MiT). mIoU is averaged across all applicable severity levels except for the noise corruption category that takes into account for the first three out of the five severity levels.}
    \label{tab:cityscape_c_comparison}
    \scalebox{0.75}{
\begin{tabular}{c|c|cccc|cccc|cccc|cccc}
\hline
\specialrule{1pt}{1pt}{1pt}
\multirow{2}{*}{Method} & \multirow{2}{*}{Clean} & \multicolumn{4}{c|}{Blur}                                     & \multicolumn{4}{c|}{Noise}                                    & \multicolumn{4}{c|}{Digital}                                  & \multicolumn{4}{c}{Weather}                                   \\ \cline{3-18} 
                        &                        & Motion        & Defoc         & Glass         & Gauss         & Gauss         & Impul         & Shot          & Speck         & Bright        & Contr         & Satur         & JPEG          & Snow          & Spat          & Fog           & Frost         \\ \hline
SegFormer               & 76.2                   & 59.3          & 59.8          & 48.7          & 60.0          & 26.2          & 27.5          & 31.1          & 52.0          & 73.2          & 66.6          & 72.0          & 38.3          & 21.5          & 53.2          & 67.2          & 31.8          \\
FeedFormer              & 77.9                   & 59.5          & 60.0          & 50.3          & 59.4          & 21.3          & 21.5          & 25.6          & 47.4          & 74.0          & 66.7          & 73.6          & 39.1          & 22.2          & 52.0          & 65.9          & 32.1          \\ \hline
\textbf{U-MixFormer}    & \textbf{79.0}          & \textbf{62.4} & \textbf{61.7} & \textbf{51.7} & \textbf{62.1} & \textbf{32.8} & \textbf{34.4} & \textbf{38.4} & \textbf{56.7} & \textbf{76.2} & \textbf{67.2} & \textbf{74.8} & \textbf{42.7} & \textbf{27.5} & \textbf{56.1} & \textbf{68.0} & \textbf{33.9} \\ \hline
\specialrule{1pt}{1pt}{1pt}
\end{tabular}
}
\end{table*}

\subsection{Experimental Results}
We compare our results with existing semantic segmentation methods on the ADE20K and Cityscapes datasets. Table \ref{tab:perform_comparison_lightmiddle} showcases our results, including the number of parameters, Floating Point Operations (FLOPs), and mIoU across both datasets. As shown in Figure~\ref{fig:fig1_performance_comparison}, we plot the performance-computation curves of different methods on the Cityscapes and ADE20K validation set.

\subsubsection{Light- and Middle-weight Models}
In Table~\ref{tab:perform_comparison_lightmiddle}, the upper section shows the performance of the light-weight models. As
shown in the table on ADE20K, our light-weight U-MixFormer-B0 establishes 41.2\% mIoU with 6.1M parameters and 6.1 GFLOPs, outperforming all other light-weight counterparts in terms of FLOPs and mIoU demonstrating a better trade-off of performance-computation. Notably, compared to SegFormer and FeedFormer, which employ the same encoder (MiT-B0), U-MixFormer achieved mIoU improvement of 3.8\% and 2.0\% while reducing computation by 27.3\% and 21.8\%. The performance disparity is even more pronounced on Cityscapes, where our model achieves 79.0\% mIoU with only 101.7 GFLOPs, indicating mIoU increases of 2.8\% and 1.1\% and computation reductions of 18.9\% and 5.3\% compared to SegFormer-B0 and FeedFormer-B0, respectively. 
When utilizing LVT, our model's performance observes further enhancements with 2.7\% and 1.3\% of increased mIoU across datasets. Furthermore, our U-MixFormer with MSCAN-T, a recent encoder from SegNeXt, also delivers standout results: 44.4\% and 81.0\% mIoU on ADE20K and Cityscapes, respectively, using 6.7M parameters. 

The latter section of Table~\ref{tab:perform_comparison_lightmiddle} shifts the focus to middle-weight models, where our approach continues to demonstrate superior results, maintaining its edge over competitors.

\subsubsection{Heavy-weight Models}
As detailed in Table~\ref{append_tab:perform_comparison_heavy}, U-MixFormer outperforms SegFormer when paired with the same heavy encoders, specifically MiT-B3/4/5. For instance, on ADE20K, U-MixFormer-B3 yields 49.8\% mIoU with only 56.8 GFLOPs. This shows a 0.4\% improvement in mIoU and a 28.1\% reduction in computation compared to SegFormer-B3. 
We additionally hypothesized that enlarging the model size (changing from MiT-B0 to MiT-B5) would allow for richer contextual information extraction from the encoder stage, potentially boosting performance. For this reason, we trained and evaluated heavy-weight model variants, MiT-B4 and MiT-B5, introducing a method to extract additional keys and values from the encoder's 3rd stage midpoint where stack lots of attention blocks. We have termed this enhanced variant, U-MixFormer+. For MiT-B4 and MiT-B5 configurations, we extracted 5 and 6 keys and values, respectively, to facilitate mix-attention. As a result, we observed a reasonable performance improvement of 0.8\% for MiT-B4 and 0.1\% for MiT-B5 when integrating more contextual data from the encoder, with only a marginal increase in computational demands.

\begin{figure}[h]
\centering
\includegraphics[width=1.0\columnwidth]{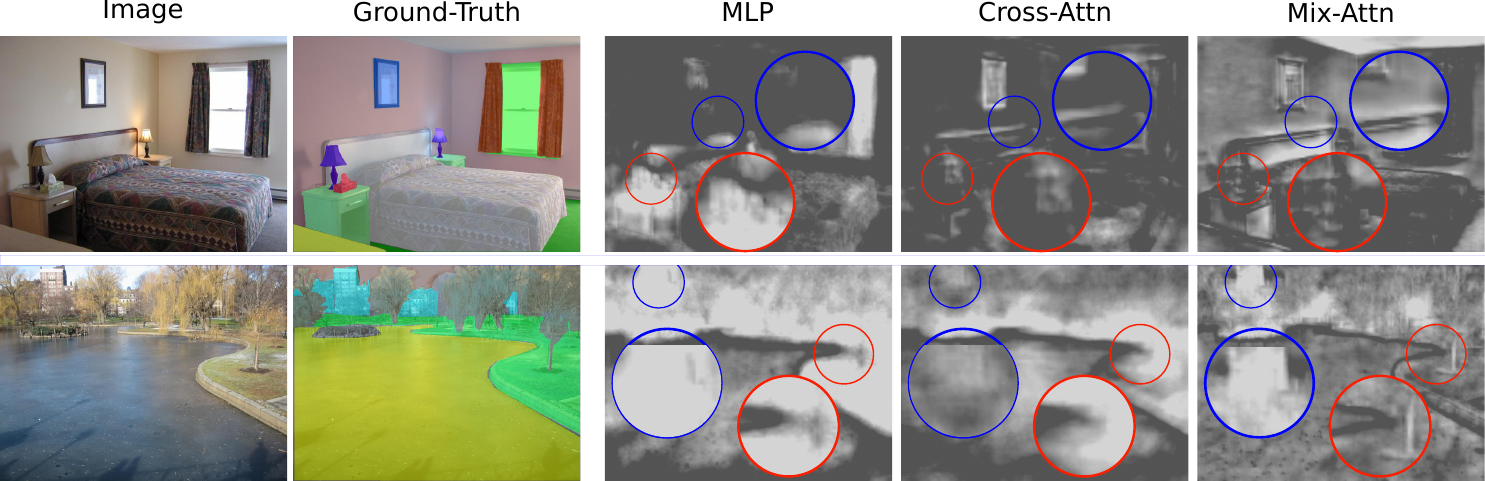}
\caption{Qualitative comparison between MLP, Cross attention, and the proposed mix-attention approach on ADE20K. We can clearly see the border segment of objects (Top row: wall/bed (blue) and ramp/box/bed (red), bottom row: building/background (blue) and lake/floor (red)), which can lead to boost semantic segmentation performance.}
\label{fig:comparison_feature_map}
\end{figure}

\subsection{Qualitative Results} 
Figure~\ref{fig:qualitative_result} presents the qualitative results of U-MixFormer, FeedFormer, and SegFormer, all utilizing identical encoders on the ADE20K and Cityscapes datasets. U-MixFormer excels in segmenting intricate object details and challenging regions more clearly than other approaches. It can significantly identify semantically relevant areas and object details, underlying its ability to learn contextual feature representations from the multi-stage encoder for efficient segmentation.

\begin{table*}[h]
    \centering
    \caption{Performance comparison with MSCAN encoder on ADE20K and Cityscapes.}
    \label{tab:perform_comparison_segnext}
    \scalebox{0.80}{
    \begin{tabular}{c|c|c|cc|cc}
    \specialrule{1pt}{1pt}{1pt}
\hline
                                  &                                    &                                       & \multicolumn{2}{c|}{\textbf{ADE20K}} & \multicolumn{2}{c}{\textbf{Cityscapes}}    \\ \cline{4-7} 
\multirow{-2}{*}{\textbf{Method}} & \multirow{-2}{*}{\textbf{Encoder}} & \multirow{-2}{*}{\textbf{Params (M)~$\downarrow$}} & \textbf{GFLOPs~$\downarrow$}    & \textbf{mIoU~$\uparrow$}   & \textbf{GFLOPs~$\downarrow$} & \textbf{mIoU~$\uparrow$}            \\ \hline
SegNeXt                           & MSCAN-T                            & 4.3                                   & 6.6                & 41.1            & 50.5            & 79.8                     \\
SegNeXt                           & MSCAN-S                            & 13.9                                  & 15.9               & 44.3            & 124.6           & 81.3                     \\
\textbf{U-MixFormer}              & \textbf{MSCAN-T}                   & \textbf{6.7}                          & \textbf{7.6}       & \textbf{44.4}   & \textbf{90.0}   & \textbf{81.0}            \\

\textbf{U-MixFormer}              & \textbf{MSCAN-S}                   & \textbf{24.3}                         & \textbf{20.8}      & \textbf{48.4}   & \textbf{154.0}  & \textbf{81.8} \\
SegNeXt                           & MSCAN-B                            & 27.6                                  & 34.9               & 48.5            & 275.7           & 82.6                    \\ \hline
\specialrule{1pt}{1pt}{1pt}
\end{tabular}
    }
\end{table*}

\subsection{Ablation Studies}
\label{ablation}
\subsubsection{Effectiveness of Mix-Attention and U-Net like Architecture}
In Table~\ref{tab:ablation_mix_unet}, we systematically evaluate model performance for various design choices. To guarantee fairness in our comparison, all models are trained and evaluated under a uniform random seed. We designate FeedFormer, which is based on conventional cross-attention, as our baseline. Integrating contextual information from multiple encoder stages via the \textit{mix-attention} module results in a 0.7\% mIoU improvement while reducing computational costs. Adapting a U-Net transformer decoder without mix-attention boosts mIoU by 0.9\% with a slight FLOPs increase(+0.4). Remarkably, by applying the mix-attention module for the U-Net architecture, the model's performance increases to 41.2\%, signifying a substantial enhancement over the traditional cross-attention in U-Net like configurations.

\subsubsection{Robustness in Image Corruption}
In safety-critical domains such as autonomous driving and intelligent transportation systems, the robustness of image segmentation is paramount. In this respect, we assessed the robustness of U-MixFormer against corruption and disturbances. Following ~\cite{City_CKamann20}, we introduced Cityscapes-C, an augmented version of the Cityscapes~\textit{val}, encompassing 16 algorithmic corruptions spanning noise, blur, weather, and digital categories. We compare our U-MixFormer with SegFormer and FeedFormer, both sharing the same encoder. The findings, shown in Table~\ref{tab:cityscape_c_comparison} highlight the superior robustness of U-MixFormer. Notably, it achieved significant improvements across all corruption categories, with a remarkable margin of up to 20.0\% and 33.3\% against shot noise and 21.8\% and 19.2\% in snowy conditions. These results demonstrate the robustness of U-MixFormer, making it an ideal choice for applications where safety and reliability are crucial.

\subsubsection{Effectiveness of Mix-Attention}
To validate the efficacy of the mix-attention module based on U-Net microscopically, we conducted an ablation study by extracting feature maps from identical locations compared to the common study (MLP and Cross-attention) in Figure~\ref{fig:comparison_feature_map}. We observed a distinct enhancement in our approach. Specifically, the feature map depicted object details with greater precision and clearly delineated the boundaries between objects. This observation demonstrates that the proposed method can significantly segment and capture distinguishable visual details locally and globally.

\subsubsection{Effectiveness of Decoding head with the same encoder}
We aimed to benchmark our methodology against the state-of-the-art using a consistent encoder, specifically adopting MiT and MSCAN. These multi-stage design encoders have received significant attention for their efficiency and innovative design. As highlighted by results in SegNeXt, both MiT and MSCAN achieved higher mean Intersection over Union (mIoU) scores and reduced computational overhead, evident from their fewer FLOPs. This analysis is vital to highlight the advantages of our method compared to these established encoders. As depicted in Table~\ref{tab:perform_comparison_lightmiddle} and ~\ref{append_tab:perform_comparison_heavy}, our U-MixFormer consistently outperforms across various model complexities from B0 to B5 in both mIoU and FLOPs. Table~\ref{tab:perform_comparison_segnext} further highlights U-MixFormer's superiority, achieving a 3.3\% mIoU increase on ADE20K over the heavier SegNeXt-S. Additionally, MSCAN-S closely rivals the performance of the more intensive MSCAN encoder (SegNeXt-B). These findings demonstrate U-MixFormer is a promising decoder architecture in semantic segmentation.

\subsubsection{Limitation and Future Works}
Despite the competitive results regarding the computational cost and mIoU of our U-MixFormer, certain limitations need to be addressed. 
We tested the inference time of a single 2048~$\times$~1024 image using a single A100 GPU under the \textit{mmsegmentation} benchmark setup.
As evident from Table~\ref{tab:perform_comparison_inference_time}, U-MixFormer tends to have a slower inference time than other light-weight models. The latency can be attributed to the inherent structure of the U-Net, which necessitates the preservation of information through lateral (or residual) connections. While essential for capturing hierarchical features, these connections introduce overheads during the inference phase. To address this limitation, we aim to explore model compression techniques such as pruning and knowledge distillation in our future work. These approaches are anticipated to potentially improve the inference speed while retaining the accuracy benefits of U-MixFormer.

\begin{table}[]
    \centering
    \caption{Comparison of inference times with light-weight models}
    \label{tab:perform_comparison_inference_time}
    \scalebox{0.80}{
\begin{tabular}{l|cccc}
    \specialrule{1pt}{1pt}{1pt}
\hline
\textbf{Method}                & \textbf{Params (M)} & \textbf{GFLOPs} & \textbf{mIoU} & \textbf{Inf. time (ms)} \\ \hline
PSPNet                         & 13.7                & 423.4           & 70.2          & 26.7                    \\
DeepLabV3+                     & 15.4                & 125.5           & 75.2          & 36.0                    \\
SegFormer                      & 3.8                 & 125.5           & 76.2          & 44.8                    \\
FeedFormer                     & 4.5                 & 107.4           & 77.9          & 54.4                    \\
\textbf{U-MixFormer}  & \textbf{6.1}        & \textbf{101.7}  & \textbf{79.0} & \textbf{55.4}           \\
\textbf{U-MixFormer} & \textbf{6.7}        & \textbf{90.0}   & \textbf{81.0} & \textbf{68.4}        \\ \hline
\specialrule{1pt}{1pt}{1pt}   
\end{tabular}
}
\end{table}
\section{Conclusion}
\label{sec:conclusion}
In this paper, we present U-MixFormer, built upon the
U-Net structure designed for semantic segmentation. U-MixFormer starts with the most contextual encoder feature map and progressively incorporates finer details, building upon U-Net's capability to capture and propagate hierarchical features. Our mix-attention design emphasizes components of merged feature maps, aligning them with increasingly granular lateral encoder features. This ensures a harmonious fusion of high-level contextual information with intricate low-level details, which is pivotal for precise segmentation. We demonstrate the superiority of our U-MixFormer across diverse encoders on popular benchmark datasets.


\bibliography{aaai24}


\end{document}